\documentclass{article}

\usepackage{microtype}
\usepackage{graphicx}
\usepackage{subcaption}
\usepackage{booktabs}
\usepackage{multirow}
\usepackage{xcolor}
\usepackage{tikz}
\usetikzlibrary{arrows.meta,positioning,shapes.geometric,fit,backgrounds,calc}
\usepackage{hyperref}

\usepackage[preprint]{icml2026}
\makeatletter\renewcommand{\ICML@preprint}{\textit{Preprint.}}\makeatother

\usepackage{amsmath}
\usepackage{amssymb}
\usepackage{mathtools}
\usepackage[capitalize,noabbrev]{cleveref}

\newcommand{\sd}[1]{\,{\scriptsize$\pm$#1}}

\icmltitlerunning{HierSentNet: Hierarchical Sentiment Analysis Framework for Hate Speech Detection}

\begin{document}

\twocolumn[
  \icmltitle{HierSentNet: Hierarchical Sentiment Analysis Framework for Hate Speech\\
    Detection Using Binary and Multiclass Classification}

  \begin{icmlauthorlist}
    \icmlauthor{Shahran Rahman Alve}{nsu}
    \icmlauthor{Faria Naznin}{nsu}
    \icmlauthor{Md Touhidur Rahman}{nsu}
    \icmlauthor{Mohammad Ashrafuzzaman Khan}{nsu}
  \end{icmlauthorlist}

  \icmlaffiliation{nsu}{Department of Electrical and Computer Engineering, North South University, Dhaka, Bangladesh}

  \icmlcorrespondingauthor{Shahran Rahman Alve}{shahran.alve@northsouth.edu}

  \icmlkeywords{hate speech detection, hierarchical classification, sentiment analysis, class imbalance, robustness}

  \vskip 0.3in
]

\printAffiliationsAndNotice{}

\begin{abstract}
Hate speech detectors must separate hateful posts from the far more common offensive ones, and two popular ideas for doing so are hierarchical classification and sentiment features. We define HierSentNet, a framework that combines a dominant-class-first hierarchy with sentiment features, and test each component in a controlled study on Davidson et al.\ (24,783 tweets) and HateXplain (19,229 posts) with five random splits and paired significance tests. With linear n-gram models, the dominant-class-first hierarchy of HierSentNet lowered macro-F1 by 1.1--1.5 points ($p\le0.007$), and VADER sentiment features never changed class-weighted Davidson macro-F1 significantly, because hateful and offensive tweets are almost equally negative. Class weighting was the only large effect (+2.0 to +3.5 macro-F1, +5.9 to +9.1 hate F1). These results held for NRC emotion features, soft hierarchical inference, smaller training sets, other imbalance remedies and a convolutional network. On OLID, whose labels are hierarchical, a matching hierarchy improved only its first decision (+1.4 macro-F1, $p=0.009$), and sentiment helped (+2.8, $p=0.005$) because there it separates offensive from non-offensive text. Robustness is the main weakness: cross-dataset transfer cut macro-F1 to 43--53, a reclaimed slur doubled the false hate rate, hate recall ranged from 54\% to 85\% across target communities, and on the HateCheck functional tests the detectors caught only 12--31\% of hateful cases. Errors concentrate where annotators disagree, and temperature scaling restores the calibration lost to class weighting.
\end{abstract}

\section{Introduction}
\label{sec:intro}

Automated moderation must flag hate speech, meaning language that attacks or demeans people because of a protected characteristic, without also flagging the much larger volume of profane, rude, or merely negative posts \citep{schmidt2017survey,davidson2017automated,fortuna2018survey,macavaney2019hate}. In the widely used dataset of \citet{davidson2017automated}, 77\% of tweets are labelled offensive and fewer than 6\% hateful, and the boundary between the two depends on targets and intent rather than on vocabulary alone. Lexicon-based detectors therefore over-flag \citep{kwok2013locate,burnap2015cyber}, and supervised detectors inherit annotation and dialect biases \citep{sap2019risk,davidson2019racial,wiegand2019detection}.

Two design ideas are often proposed to help with this boundary. The first is \emph{hierarchical} classification \citep{koller1997hierarchically,silla2011survey}: separate a coarse distinction first, then refine it, as in the OLID annotation scheme \citep{zampieri2019predicting}. The second is \emph{affective} information: hateful messages tend to carry negative sentiment and emotions such as anger and disgust, so sentiment or emotion signals have been added as features or auxiliary tasks \citep{samghabadi2019attending,rajamanickam2020joint,plazadelarco2021multitask}. A natural way to combine both ideas, which we call HierSentNet, first separates the dominant class from a merged minority class, then classifies the minority classes, and adds sentiment features at every stage. Designs of this kind are attractive because each stage faces a simpler and more balanced decision, but the contribution of each component is rarely isolated, and strong encoders such as RoBERTa \citep{liu2019roberta} and XLM-RoBERTa \citep{conneau2020xlmr} make it easy to attribute gains to the architecture rather than to the training choices around it.

This paper asks whether each component actually helps, under a protocol strict enough to detect small effects:

\begin{itemize}
  \item \textbf{RQ1.} Does a two-stage hierarchy outperform a flat three-way classifier, and does the choice of the first split matter?
  \item \textbf{RQ2.} Do sentiment features improve hate detection, and why or why not?
  \item \textbf{RQ3.} How much of the gain comes from simply handling class imbalance?
  \item \textbf{RQ4.} How robust are these detectors to a change of dataset and to dialect-associated vocabulary?
  \item \textbf{RQ5.} Do the conclusions survive stronger variants: emotion categories instead of polarity, probabilistic instead of hard hierarchies, other imbalance remedies, less training data, and a neural model?
  \item \textbf{RQ6.} Do they hold on a hierarchically annotated dataset (OLID) and under functional tests (HateCheck), and how much do annotator disagreement and calibration matter?
\end{itemize}

\textbf{Contributions.}
\begin{itemize}
  \setlength\itemsep{1pt}
  \item We formalise two hierarchical decompositions of the hate/offensive/neither task, including the dominant-class-first scheme of HierSentNet.
  \item We run a controlled study of 24 configurations (two linear learners $\times$ three decompositions $\times$ sentiment on/off $\times$ class weighting on/off) on two benchmarks, with five seeded splits, validation-tuned regularisation and paired bootstrap tests.
  \item We show that the dominant-first hierarchy hurts, sentiment features are neutral on the hate/offensive boundary, and class weighting dominates, and we explain the sentiment result with the sentiment distributions of the classes and the hierarchy result with confusion matrices.
  \item We quantify cross-dataset degradation and a lexical false-positive bias.
  \item We stress-test every conclusion with emotion-category features, probabilistic hierarchical inference, learning curves, five imbalance remedies with a calibration analysis, a neural model, and a per-target error analysis.
  \item We add a hierarchically annotated benchmark (OLID), functional tests (HateCheck), an annotator-agreement analysis, a representation ablation and temperature scaling.
  \item We release code, including a GPU script that runs transformer models on exactly the same splits.
\end{itemize}
\Cref{fig:workflow} summarises the study design.

\begin{figure*}[t]
\centering
\definecolor{cdata}{HTML}{4C72B0}\definecolor{cfeat}{HTML}{DD8452}\definecolor{cprot}{HTML}{55A868}\definecolor{cmod}{HTML}{8172B3}\definecolor{cleak}{HTML}{C44E52}
\begin{tikzpicture}[font=\scriptsize, node distance=1.4mm and 4mm,
  box/.style={draw, rounded corners=2pt, align=center, inner sep=2pt, minimum height=5.5mm, text width=27mm},
  hdr/.style={font=\scriptsize\bfseries, align=center},
  grp/.style={draw, rounded corners=4pt, inner sep=1.5mm},
  arr/.style={-{Stealth[length=1.4mm]}, thin}]
\node[hdr] (h1) at (0mm,0) {Data};
\node[box, fill=cdata!12, below=1.2mm of h1] (d1) {\textbf{Davidson}: 24,783 tweets\\hate 6\%, offensive 77\%};
\node[box, fill=cdata!12, below=of d1] (d2) {\textbf{HateXplain}: 19,229 posts\\hate is the 2nd largest class};
\node[box, fill=cdata!12, below=of d2] (d3) {Cleaning: URLs, mentions,\\RT, elongation, lower case};
\node[box, fill=cdata!12, below=of d3] (d4) {5 stratified 80/10/10 splits;\\official HateXplain split};
\begin{scope}[on background layer]\node[grp, draw=cdata, fit=(h1)(d1)(d4)] (g1) {};\end{scope}
\node[hdr] (h2) at (34.5mm,0) {Representation};
\node[box, fill=cfeat!12, below=1.2mm of h2] (f1) {TF-IDF word 1--2-grams\\+ character 2--5-grams};
\node[box, fill=cfeat!12, below=of f1] (f2) {\textbf{Polarity}: VADER\\neg, neu, pos, compound};
\node[box, fill=cfeat!12, below=of f2] (f3) {\textbf{Emotion}: NRC lexicon,\\10 categories + coverage};
\node[box, fill=cfeat!12, below=of f3] (f4) {Learned embeddings\\(TextCNN, from scratch)};
\begin{scope}[on background layer]\node[grp, draw=cfeat, fit=(h2)(f1)(f4)] (g2) {};\end{scope}
\node[hdr] (h3) at (69mm,0) {Design choices};
\node[box, fill=cprot!12, below=1.2mm of h3] (p1) {\textbf{Decomposition}: flat,\\dominant-first, abusive-first};
\node[box, fill=cprot!12, below=of p1] (p2) {\textbf{Inference}: hard\\or soft (probabilistic)};
\node[box, fill=cprot!12, below=of p2] (p3) {\textbf{Imbalance}: weights, over-/\\undersampling, decision bias};
\node[box, fill=cprot!12, below=of p3] (p4) {\textbf{Training size}:\\10, 25, 50, 100\%};
\begin{scope}[on background layer]\node[grp, draw=cprot, fit=(h3)(p1)(p4)] (g3) {};\end{scope}
\node[hdr] (h4) at (103.5mm,0) {Learners};
\node[box, fill=cmod!12, below=1.2mm of h4] (m1) {\textbf{LogReg} (one-vs-rest,\\LIBLINEAR)};
\node[box, fill=cmod!12, below=of m1] (m2) {\textbf{Linear SVM}\\(squared hinge)};
\node[box, fill=cmod!12, below=of m2] (m3) {\textbf{TextCNN}\\(filters 3, 4, 5)};
\node[box, fill=cmod!12, below=of m3] (m4) {$C$ / epoch chosen on\\validation macro-F1};
\begin{scope}[on background layer]\node[grp, draw=cmod, fit=(h4)(m1)(m4)] (g4) {};\end{scope}
\node[hdr] (h5) at (138mm,0) {Evaluation};
\node[box, fill=gray!10, below=1.2mm of h5] (e1) {Macro-F1, hate P/R/F1,\\\textsc{o}$\to$\textsc{h} errors, ECE};
\node[box, fill=gray!10, below=of e1] (e2) {Paired bootstrap;\\paired $t$-test over splits};
\node[box, fill=cleak!10, draw=cleak, below=of e2] (e3) {Cross-dataset transfer;\\reclaimed-slur bias probe};
\node[box, fill=cleak!10, draw=cleak, below=of e3] (e4) {Hate recall per\\target community};
\begin{scope}[on background layer]\node[grp, draw=black!60, fit=(h5)(e1)(e4)] (g5) {};\end{scope}
\draw[arr, thick] (g1.east) -- (g2.west); \draw[arr, thick] (g2.east) -- (g3.west); \draw[arr, thick] (g3.east) -- (g4.west); \draw[arr, thick] (g4.east) -- (g5.west);
\end{tikzpicture}
\vspace{-1mm}
\caption{\textbf{Workflow of the controlled study.} Two public benchmarks are cleaned and split identically for every configuration. Each model combines a text representation with optional polarity or emotion features, a decomposition and inference rule, an imbalance remedy and a learner, and is evaluated with per-class metrics, significance tests and robustness probes (red). Every design choice is varied while the others are held fixed, so each effect can be measured in isolation.}
\label{fig:workflow}
\vspace{-2mm}
\end{figure*}

\section{Related Work}
\label{sec:related}

\textbf{Definitions, surveys and datasets.} Hate speech has no single accepted definition, and surveys stress that this ambiguity propagates into datasets and models \citep{schmidt2017survey,fortuna2018survey,macavaney2019hate,yin2021towards}. Early English corpora collected tweets with keyword or hashtag searches and labelled them for racism and sexism \citep{waseem2016hateful} or for hate versus offence \citep{davidson2017automated}. Later resources grew in size and scope: large crowdsourced abuse corpora \citep{founta2018large}; shared tasks on hate against immigrants and women \citep{basile2019semeval} and on offensive language in several languages \citep{zampieri2019predicting,zampieri2020semeval}; HateXplain, with token-level rationales and target labels \citep{mathew2021hatexplain}; implicit hate \citep{elsherief2021latent}; social-bias inference \citep{sap2020social}; and adversarially collected data \citep{vidgen2021learning}. \citet{vidgen2020directions} review over sixty such datasets and show how sampling and labelling choices shape what models learn.

\textbf{Models for detection.} Feature-based systems combined word and character n-grams, lexicons and linear classifiers \citep{warner2012detecting,kwok2013locate,burnap2015cyber,waseem2016hateful,davidson2017automated}. Neural models followed, including CNNs \citep{gamback2017using} and LSTMs with gradient-boosted trees \citep{badjatiya2017deep}. Pretrained transformers \citep{devlin2019bert,liu2019roberta} now dominate: BERT fine-tuning transfers well to hate speech \citep{mozafari2019bert}, domain-adaptive pretraining on abusive Reddit text yields HateBERT \citep{caselli2021hatebert}, multi-channel BERT combines languages and translations \citep{sohn2019mcbert4hate}, and XLM-R extends the approach to many languages \citep{conneau2020xlmr}. Unified benchmarks such as TweetEval \citep{barbieri2020tweeteval} made comparisons more systematic. Strong linear n-gram models remain competitive baselines on these datasets, which is why we use them to isolate design choices.

\textbf{Affect-aware detection.} Basic-emotion theory \citep{ekman1992argument} and affect datasets such as SemEval-2018 Affect in Tweets \citep{mohammad2018semeval} motivate using sentiment and emotion as cues. Emotion has been used as attention guidance \citep{samghabadi2019attending}, as an auxiliary task in multi-task learning \citep{caruana1997multitask,rajamanickam2020joint,plazadelarco2021multitask}, jointly with target detection \citep{awal2021angrybert}, and across different hate targets \citep{chiril2022emotionally}. These works report gains, mostly from emotion categories rather than polarity. Our study tests the simpler polarity-feature variant, then emotion categories, and explains why neither helps on the hate/offensive boundary.

\textbf{Hierarchical classification.} Top-down classifiers decompose a label set into a tree of easier decisions, which helps when the tree matches the structure of the data \citep{koller1997hierarchically,dumais2000hierarchical,silla2011survey}; neural variants include HDLTex \citep{kowsari2017hdltex} and hierarchical multi-label networks \citep{wehrmann2018hierarchical}. In abusive language, OLID annotates offensiveness, then type, then target \citep{zampieri2019predicting}. A known weakness of top-down inference is error propagation: a mistake at the first level cannot be corrected below \citep{silla2011survey}. We test two orders of the same three classes to measure this effect directly.

\textbf{Class imbalance.} Standard remedies include resampling such as SMOTE \citep{chawla2002smote}, cost-sensitive learning \citep{he2009learning}, and loss reweighting by inverse or effective class frequency or by difficulty \citep{lin2017focal,cui2019classbalanced}. Hate speech data are strongly imbalanced, yet many detection papers do not report how imbalance is handled, which confounds comparisons between architectures. We treat class weighting as an explicit experimental factor.

\textbf{Robustness, bias and annotation.} Hate speech models generalise poorly across datasets \citep{arango2019hate,swamy2019studying,yin2021towards} and can exploit dataset artefacts \citep{gururangan2018annotation,wiegand2019detection}. They show unintended bias against identity terms \citep{dixon2018measuring,kennedy2020contextualizing} and flag African-American English more often \citep{sap2019risk,davidson2019racial}. Behavioural test suites such as CheckList \citep{ribeiro2020beyond} and HateCheck \citep{rottger2021hatecheck} expose failures that aggregate metrics hide, and majority-vote labels hide genuine annotator disagreement \citep{davani2022dealing}. Following this line of work, we report cross-dataset transfer and a lexical bias probe alongside in-domain scores.

\section{Problem Setup and Methods}
\label{sec:method}

\textbf{Task.} Given a post $x$, predict $y\in\{\textsc{h},\textsc{o},\textsc{n}\}$ (hate, offensive, neither). Because \textsc{o} dominates, we evaluate with macro-F1 and the F1 of the hate class.

\textbf{Flat classifier.} A single classifier $f(x)$ over the three classes.

\textbf{Dominant-first hierarchy (HierSentNet).} Stage 1 learns $g_1(x)$ to separate the dominant class \textsc{o} from a merged minority class $\{\textsc{h},\textsc{n}\}$. Stage 2 learns $g_2(x)$ to separate \textsc{h} from \textsc{n}, trained only on minority examples:
\begin{equation}
\hat y(x)=\begin{cases}\textsc{o} & \text{if } g_1(x)=\textsc{o},\\ g_2(x)\in\{\textsc{h},\textsc{n}\} & \text{otherwise.}\end{cases}
\label{eq:dom}
\end{equation}
This merge-then-split design is the decomposition used in HierSentNet: stage 1 separates the dominant label from all other labels.

\textbf{Abusive-first hierarchy.} Stage 1 separates abusive content $\{\textsc{h},\textsc{o}\}$ from \textsc{n}; stage 2 separates \textsc{h} from \textsc{o}. This follows the OLID-style order \citep{zampieri2019predicting}, in which the hard hate/offensive decision is made only on abusive posts.

\begin{figure*}[t]
\centering
\definecolor{cH}{HTML}{C44E52}\definecolor{cO}{HTML}{DD8452}\definecolor{cN}{HTML}{4C72B0}\definecolor{cS}{HTML}{8172B3}
\begin{tikzpicture}[font=\scriptsize,
  inp/.style={draw, rounded corners=2pt, fill=gray!10, align=center, inner sep=2pt, minimum height=5mm, minimum width=14mm},
  dec/.style={draw=cS, thick, rounded corners=2pt, fill=cS!10, align=center, inner sep=2pt, minimum height=5.5mm, minimum width=22mm},
  lH/.style={draw=cH, rounded corners=5pt, fill=cH!15, minimum width=7mm, minimum height=4.5mm},
  lO/.style={draw=cO, rounded corners=5pt, fill=cO!15, minimum width=7mm, minimum height=4.5mm},
  lN/.style={draw=cN, rounded corners=5pt, fill=cN!15, minimum width=7mm, minimum height=4.5mm},
  hdr/.style={font=\scriptsize\bfseries},
  note/.style={font=\tiny, align=center, text width=40mm},
  arr/.style={-{Stealth[length=1.3mm]}, thin},
  grp/.style={draw=black!35, rounded corners=4pt, inner sep=1.8mm}]
\node[hdr] (ta) at (15mm,0) {(a) Flat};
\node[inp] (xa) at (15mm,-7mm) {post $x$};
\node[dec] (fa) at (15mm,-16mm) {$f$: \textsc{h} / \textsc{o} / \textsc{n}};
\node[lH] (aH) at (4mm,-27mm) {\textsc{h}}; \node[lO] (aO) at (15mm,-27mm) {\textsc{o}}; \node[lN] (aN) at (26mm,-27mm) {\textsc{n}};
\draw[arr] (xa) -- (fa); \foreach \t in {aH,aO,aN} \draw[arr] (fa) -- (\t);
\node[note, text width=28mm] (na) at (15mm,-44mm) {one decision;\\every class is seen in training};
\begin{scope}[on background layer]\node[grp, fit=(ta)(aH)(aN)(na)(fa)] {};\end{scope}
\node[hdr] (tb) at (72mm,0) {(b) Dominant-first};
\node[inp] (xb) at (72mm,-7mm) {post $x$};
\node[dec] (g1b) at (72mm,-16mm) {$g_1$: \textsc{o} vs.\ \{\textsc{h},\textsc{n}\}};
\node[lO] (bO) at (56mm,-27mm) {\textsc{o}};
\node[dec] (g2b) at (82mm,-27mm) {$g_2$: \textsc{h} vs.\ \textsc{n}};
\node[lH] (bH) at (74mm,-37mm) {\textsc{h}}; \node[lN] (bN) at (90mm,-37mm) {\textsc{n}};
\draw[arr] (xb) -- (g1b); \draw[arr] (g1b) -- (bO); \draw[arr] (g1b) -- (g2b); \draw[arr] (g2b) -- (bH); \draw[arr] (g2b) -- (bN);
\node[note, text=cH] (nb) at (72mm,-45mm) {$g_2$ never sees an \textsc{o} example, so an \textsc{o} post that passes $g_1$ must become \textsc{h} or \textsc{n}};
\begin{scope}[on background layer]\node[grp, fit=(tb)(bO)(bN)(nb)] {};\end{scope}
\node[hdr] (tc) at (136mm,0) {(c) Abusive-first};
\node[inp] (xc) at (136mm,-7mm) {post $x$};
\node[dec] (g1c) at (136mm,-16mm) {$g_1$: \{\textsc{h},\textsc{o}\} vs.\ \textsc{n}};
\node[lN] (cN) at (152mm,-27mm) {\textsc{n}};
\node[dec] (g2c) at (126mm,-27mm) {$g_2$: \textsc{h} vs.\ \textsc{o}};
\node[lH] (cH) at (118mm,-37mm) {\textsc{h}}; \node[lO] (cO) at (134mm,-37mm) {\textsc{o}};
\draw[arr] (xc) -- (g1c); \draw[arr] (g1c) -- (cN); \draw[arr] (g1c) -- (g2c); \draw[arr] (g2c) -- (cH); \draw[arr] (g2c) -- (cO);
\node[note] (nc) at (136mm,-45mm) {the hard \textsc{h}/\textsc{o} decision is made only on posts already judged abusive};
\begin{scope}[on background layer]\node[grp, fit=(tc)(cH)(cN)(nc)] {};\end{scope}
\node[draw=cS, dashed, rounded corners=3pt, fill=cS!5, align=center, inner sep=3pt, font=\tiny] (soft) at (78mm,-56mm)
 {\textbf{Soft (probabilistic) inference}, shown for (c):\quad $P(\textsc{h})=P_1(\mathrm{ab})\,P_2(\textsc{h}|\mathrm{ab})$,\quad $P(\textsc{o})=P_1(\mathrm{ab})\,P_2(\textsc{o}|\mathrm{ab})$,\quad $P(\textsc{n})=1-P_1(\mathrm{ab})$\\[1pt]
 where ab = abusive; predict the most probable class, so a confident $g_2$ can overrule an uncertain $g_1$};
\end{tikzpicture}
\vspace{-1mm}
\caption{\textbf{The three decompositions of the hate (\textsc{h}) / offensive (\textsc{o}) / neither (\textsc{n}) task.} (a) A single flat classifier. (b) The merge-then-split design of HierSentNet: stage 1 separates the dominant offensive class from a merged minority class, and stage 2 never sees an offensive example. (c) The OLID-style order. Both hierarchies are evaluated with hard decisions and with probabilistic (soft) inference (bottom); the dominant-first version is analogous.}
\label{fig:hier}
\vspace{-2mm}
\end{figure*}

\textbf{Sentiment features.} For each post we compute the VADER scores \citep{hutto2014vader} (negative, neutral, positive, compound and $|$compound$|$), standardise them on the training split, scale them by 0.5, and append them to the text features. In a transformer, the same features can be concatenated with the \texttt{[CLS]} vector, which the released script supports.

\textbf{Emotion features.} To test whether richer affect helps where polarity does not, we also compute NRC emotion features \citep{mohammad2013crowdsourcing}: for each of the ten NRC categories (anger, anticipation, disgust, fear, joy, negative, positive, sadness, surprise, trust) the share of tokens associated with it, plus the share of tokens covered by the lexicon. They are standardised and scaled like the VADER features.

\textbf{Soft hierarchical inference.} The hierarchies above make a hard decision at stage 1. To separate the effect of the decomposition from that of error propagation, we also evaluate a probabilistic version that combines the calibrated stage probabilities. For the abusive-first order, $P(\textsc{h}\mid x)=P(\text{abusive}\mid x)\,P(\textsc{h}\mid \text{abusive},x)$, $P(\textsc{o}\mid x)=P(\text{abusive}\mid x)\,P(\textsc{o}\mid \text{abusive},x)$ and $P(\textsc{n}\mid x)=1-P(\text{abusive}\mid x)$, and analogously for the dominant-first order; the prediction is the most probable class, so a confident stage 2 can overrule an uncertain stage 1.

\begin{figure}[t]
\centering
\definecolor{cfeat}{HTML}{DD8452}\definecolor{cmod}{HTML}{8172B3}\definecolor{cdata}{HTML}{4C72B0}\definecolor{caff}{HTML}{55A868}
\begin{tikzpicture}[font=\scriptsize,
  box/.style={draw, rounded corners=2pt, align=center, inner sep=2pt, minimum height=6mm},
  arr/.style={-{Stealth[length=1.3mm]}, thin},
  lab/.style={font=\tiny\bfseries, text=black!65}]
\node[box, fill=cdata!12, minimum width=24mm] (x) at (40mm,0) {post $x$ (cleaned)};
\node[box, fill=cfeat!12, text width=14mm] (w) at (11mm,-13mm) {word\\1--2-grams};
\node[box, fill=cfeat!12, text width=14mm] (c) at (29mm,-13mm) {char\\2--5-grams};
\node[box, fill=cfeat!6, text width=32mm] (tf) at (20mm,-24mm) {TF-IDF, sublinear tf\\($\approx$80k sparse dims)};
\node[box, fill=caff!12, text width=12mm] (v) at (53mm,-13mm) {VADER\\(5 scores)};
\node[box, fill=caff!12, text width=12mm] (e) at (69mm,-13mm) {NRC\\(11 scores)};
\node[box, fill=caff!6, text width=29mm] (sc) at (61mm,-24mm) {standardise, $\times\,0.5$\\(train-split statistics)};
\node[lab] (lt) at (20mm,-30mm) {text branch};
\node[lab] (la) at (61mm,-30mm) {affect branch (optional)};
\begin{scope}[on background layer]
\node[draw=cfeat, rounded corners=3pt, inner sep=1.6mm, fit=(w)(c)(tf)(lt)] (gt) {};
\node[draw=caff, dashed, rounded corners=3pt, inner sep=1.6mm, fit=(v)(e)(sc)(la)] (ga) {};
\end{scope}
\node[box, fill=cmod!8, text width=52mm] (cat) at (40mm,-41mm) {concatenate: $\phi(x)=[\,\phi_{\text{text}}(x)\;\|\;\phi_{\text{affect}}(x)\,]$};
\node[box, fill=cmod!16, text width=52mm] (clf) at (40mm,-51mm) {linear classifier for each stage\\(class-weighted or bias-corrected)};
\node[box, fill=cmod!26, text width=52mm] (dec) at (40mm,-61.5mm) {flat, dominant-first or abusive-first, hard or soft\\\mbox{prediction $\hat y\in\{\textsc{h},\textsc{o},\textsc{n}\}$}};
\coordinate (bus) at (40mm,-5.5mm);
\draw[thin] (x.south) -- (bus);
\foreach \n in {w,c,v,e} \draw[arr] (bus) -| (\n.north);
\draw[arr] (w.south) -- (w.south |- tf.north); \draw[arr] (c.south) -- (c.south |- tf.north);
\draw[arr] (v.south) -- (v.south |- sc.north); \draw[arr] (e.south) -- (e.south |- sc.north);
\draw[arr] (gt.south -| tf.south) -- (tf.south |- cat.north); \draw[arr] (ga.south -| sc.south) -- (sc.south |- cat.north);
\draw[arr] (cat) -- (clf); \draw[arr] (clf) -- (dec);
\end{tikzpicture}
\vspace{-1mm}
\caption{\textbf{Feature fusion in the linear models.} Word and character n-grams form the text representation. VADER polarity and NRC emotion scores form an optional affect branch, standardised on the training split and down-weighted so they cannot dominate the sparse text features. The fused vector feeds every stage of a decomposition.}
\label{fig:arch}
\vspace{-2mm}
\end{figure}

\textbf{Text representation.} HTML entities are unescaped; URLs and user mentions are replaced by placeholder tokens; the retweet marker is removed; characters repeated more than three times are shortened to three; text is lower-cased. We use TF-IDF over word 1--2-grams (minimum document frequency 2, at most 100k features) and character 2--5-grams within word boundaries (minimum document frequency 3, at most 200k features), both with sublinear term frequency and fitted on the training split only (about 80k features in total on Davidson).

\Cref{fig:arch} shows how the text and affect features are combined.

\textbf{Learners.} We train one-vs-rest logistic regression (LogReg) or a linear SVM with squared hinge loss, both with LIBLINEAR \citep{fan2008liblinear} through scikit-learn \citep{pedregosa2011scikit}. Class weighting, when used, sets each class weight inversely proportional to its training frequency \citep{he2009learning}. Every stage of a hierarchy uses the same learner, features and regularisation. To check that our conclusions are not specific to linear models, we also train a convolutional network \citep{kim2014convolutional} from scratch: 128-dimensional word embeddings, filters of width 3, 4 and 5 (100 each) with max pooling, dropout 0.5, Adam with learning rate $10^{-3}$, batch size 64, and the epoch (at most five) with the best validation macro-F1.

\section{Experimental Setup}
\label{sec:setup}

\textbf{Datasets.} \emph{Davidson} \citep{davidson2017automated}: 24,783 tweets labelled hate (1,430), offensive (19,190) or neither (4,163) by majority vote of CrowdFlower annotators. \emph{HateXplain} \citep{mathew2021hatexplain}: Twitter and Gab posts labelled hatespeech, offensive or normal by three annotators; we keep the 19,229 posts with a majority label, map them to \textsc{h}/\textsc{o}/\textsc{n}, and use the official split (15,383 / 1,922 / 1,924). The two datasets differ sharply in class balance: hate is the smallest Davidson class but the second largest HateXplain class.

\textbf{Protocol.} For Davidson we draw five stratified 80/10/10 train/validation/test splits (seeds 0--4; 2,479 test tweets each). The regularisation strength $C\in\{0.3,1,3,10\}$ (the SVM uses $C/10$) is chosen per configuration on validation macro-F1 of the full pipeline, and all stages of a hierarchy share it. We report the mean and standard deviation over seeds. For significance we pool the five test sets (12,395 predictions per configuration) and run a paired bootstrap with 2,000 resamples \citep{koehn2004statistical}, reporting the difference, its 95\% interval and a two-sided $p$-value. HateXplain results use the single official split.

\textbf{Additional analyses.} (i) A lexicon baseline that flags a tweet as hateful if it contains any of the 178 n-grams of the refined hate lexicon released with the Davidson dataset. (ii) Cross-dataset transfer: train on one dataset (validation on its own split) and test on the other, with LogReg and class weighting. (iii) A lexical bias probe: the rate at which non-hateful tweets containing a reclaimed slur, written \emph{n*gga} here, are classified as hate, compared with non-hateful tweets without it \citep{sap2019risk,davidson2019racial}. (iv) Row-normalised confusion matrices for each decomposition. (v) Five strategies for class imbalance: class weights, random oversampling of the minority classes, random undersampling of the majority classes, a decision bias (per-class offsets added to the log-probabilities, tuned on validation macro-F1), and weights plus bias. (vi) Learning curves with 10, 25 and 50\% of the training data. (vii) Calibration, measured by the expected calibration error over 15 confidence bins \citep{guo2017calibration}. (viii) Hate recall for each target community annotated in HateXplain. For the extended analyses we compare variants with a paired $t$-test over the five Davidson splits, which is more conservative than the pooled bootstrap.

\textbf{Transformer protocol.} Pretrained transformer weights were not reachable in our compute environment (a 2-core CPU), so we do not report transformer results. The released script fine-tunes RoBERTa-base and XLM-RoBERTa-base on exactly the splits used here (maximum length 96, batch 32, AdamW with learning rate $2\times10^{-5}$ and weight decay 0.01, 6\% linear warm-up, 3 epochs, class-weighted cross-entropy, model selection on validation macro-F1, and optional concatenation of the VADER features to the \texttt{[CLS]} vector).

\section{Results}
\label{sec:results}

\begin{table*}[t]
\caption{All 24 configurations on both datasets (\%). Davidson: mean\,$\pm$\,std over five seeded splits; \textsc{o}$\to$\textsc{h} is the share of offensive tweets misclassified as hate. HateXplain: official test split. ``Wt.'' marks class-weighted training; ``Sent.'' marks VADER features. The lexicon row flags any tweet containing a hate-lexicon n-gram.}
\label{tab:main}
\centering
\footnotesize
\setlength{\tabcolsep}{3.4pt}
\resizebox{\textwidth}{!}{%
\begin{tabular}{@{}cllcccccccc|cccc@{}}
\toprule
 & & & & \multicolumn{7}{c|}{Davidson (5 splits)} & \multicolumn{4}{c}{HateXplain} \\
\cmidrule(lr){5-11}\cmidrule(l){12-15}
 & Wt. & Decomposition & Sent. & Macro-F1 & Hate P & Hate R & Hate F1 & Off.\ F1 & Neither F1 & \textsc{o}$\to$\textsc{h} & Macro-F1 & Hate F1 & Off.\ F1 & Normal F1 \\
\midrule
\multirow{12}{*}{\rotatebox{90}{LogReg}} & -- & Flat & -- & 71.7\sd{2.0} & 54.7 & 26.6 & 35.7\sd{6.8} & 94.2 & 85.3 & 1.5 & 65.9 & 75.7 & 48.6 & 73.5 \\
 & -- & Flat & \checkmark & 71.7\sd{2.2} & 53.0 & 26.4 & 35.2\sd{7.8} & 94.2 & 85.8 & 1.6 & 66.0 & 75.7 & 48.6 & 73.6 \\
 & -- & Dom.-first & -- & 69.9\sd{2.1} & 49.0 & 26.3 & 33.9\sd{6.4} & 93.5 & 82.2 & 1.9 & 63.1 & 74.5 & 42.2 & 72.5 \\
 & -- & Dom.-first & \checkmark & 70.2\sd{1.1} & 47.9 & 27.7 & 34.9\sd{5.2} & 93.4 & 82.4 & 2.0 & 63.2 & 75.3 & 41.9 & 72.6 \\
 & -- & Abus.-first & -- & 71.9\sd{2.1} & 57.3 & 25.3 & 35.0\sd{6.9} & 94.3 & 86.3 & 1.3 & 64.9 & 72.1 & 52.9 & 69.6 \\
 & -- & Abus.-first & \checkmark & 72.0\sd{1.9} & 57.9 & 25.5 & 35.3\sd{6.5} & 94.3 & 86.5 & 1.3 & 65.4 & 72.9 & 53.6 & 69.6 \\
\cmidrule(l){2-15}
 & \checkmark & Flat & -- & 74.5\sd{1.5} & 40.9 & 48.5 & 44.2\sd{5.3} & 93.1 & 86.2 & 5.0 & 66.2 & 73.7 & 52.6 & 72.4 \\
 & \checkmark & Flat & \checkmark & 74.4\sd{1.2} & 39.8 & 50.3 & 44.4\sd{4.5} & 93.0 & 86.0 & 5.3 & 67.0 & 74.9 & 53.3 & 72.8 \\
 & \checkmark & Dom.-first & -- & 73.5\sd{1.7} & 38.1 & 48.5 & 42.6\sd{5.5} & 92.8 & 85.0 & 5.3 & 64.8 & 71.4 & 53.9 & 69.2 \\
 & \checkmark & Dom.-first & \checkmark & 72.9\sd{2.1} & 37.4 & 46.3 & 41.3\sd{6.1} & 92.8 & 84.5 & 5.1 & 65.0 & 71.4 & 53.7 & 69.9 \\
 & \checkmark & Abus.-first & -- & 74.6\sd{1.6} & 39.8 & 46.7 & 42.9\sd{5.3} & 93.4 & 87.6 & 5.1 & 64.9 & 72.7 & 48.7 & 73.3 \\
 & \checkmark & Abus.-first & \checkmark & 74.3\sd{1.1} & 39.8 & 45.3 & 42.3\sd{4.0} & 93.4 & 87.2 & 4.9 & 66.8 & 75.1 & 51.3 & 73.9 \\
\midrule
\multirow{12}{*}{\rotatebox{90}{Linear SVM}} & -- & Flat & -- & 71.9\sd{1.6} & 50.3 & 28.0 & 35.9\sd{5.4} & 94.1 & 85.8 & 1.9 & 65.8 & 76.0 & 48.3 & 73.2 \\
 & -- & Flat & \checkmark & 72.3\sd{1.7} & 51.0 & 28.4 & 36.4\sd{5.8} & 94.2 & 86.4 & 1.9 & 65.6 & 75.4 & 48.0 & 73.4 \\
 & -- & Dom.-first & -- & 70.7\sd{1.5} & 45.6 & 29.7 & 35.8\sd{4.6} & 93.4 & 83.0 & 2.5 & 63.2 & 74.3 & 42.5 & 72.6 \\
 & -- & Dom.-first & \checkmark & 70.7\sd{1.6} & 46.1 & 28.7 & 35.0\sd{5.7} & 93.5 & 83.7 & 2.3 & 63.7 & 75.4 & 42.8 & 72.8 \\
 & -- & Abus.-first & -- & 72.6\sd{2.1} & 54.4 & 27.3 & 36.3\sd{6.5} & 94.4 & 87.1 & 1.5 & 66.1 & 73.5 & 52.7 & 72.1 \\
 & -- & Abus.-first & \checkmark & 72.6\sd{1.9} & 54.0 & 27.6 & 36.4\sd{6.1} & 94.4 & 86.9 & 1.6 & 66.8 & 74.2 & 54.0 & 72.1 \\
\cmidrule(l){2-15}
 & \checkmark & Flat & -- & 75.4\sd{1.1} & 45.7 & 44.3 & 44.9\sd{4.4} & 94.0 & 87.2 & 3.7 & 65.7 & 74.4 & 50.0 & 72.7 \\
 & \checkmark & Flat & \checkmark & 75.4\sd{1.2} & 46.4 & 44.5 & 45.4\sd{4.3} & 93.9 & 86.7 & 3.6 & 66.2 & 75.0 & 50.6 & 73.1 \\
 & \checkmark & Dom.-first & -- & 73.9\sd{1.8} & 39.6 & 48.1 & 43.4\sd{5.5} & 92.9 & 85.4 & 5.0 & 64.9 & 71.5 & 54.1 & 69.2 \\
 & \checkmark & Dom.-first & \checkmark & 73.6\sd{1.1} & 37.9 & 50.1 & 43.1\sd{3.3} & 92.8 & 84.9 & 5.4 & 65.1 & 72.2 & 53.5 & 69.7 \\
 & \checkmark & Abus.-first & -- & 74.6\sd{1.4} & 38.9 & 46.2 & 42.1\sd{4.7} & 93.4 & 88.2 & 5.2 & 66.3 & 74.1 & 51.3 & 73.5 \\
 & \checkmark & Abus.-first & \checkmark & 74.4\sd{1.6} & 38.4 & 47.0 & 42.1\sd{5.9} & 93.3 & 87.8 & 5.4 & 65.6 & 74.3 & 48.9 & 73.5 \\
\midrule
\multicolumn{3}{@{}l}{Lexicon (hate n-gram match)} & -- & -- & 47.1 & 43.5 & 45.2\sd{3.4} & -- & -- & 3.5 & -- & -- & -- & -- \\
\bottomrule
\end{tabular}}
\end{table*}

\subsection{Does the Hierarchy Help? (RQ1)}

\begin{figure}[t]
\centering
\includegraphics[width=\columnwidth]{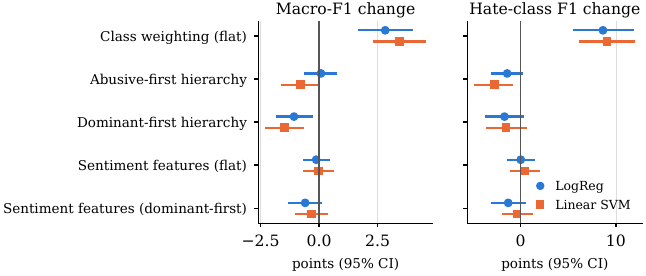}
\caption{Effect of each design choice on the pooled Davidson test predictions (paired bootstrap, 2,000 resamples, 95\% intervals). Hierarchy and sentiment rows compare against the flat, sentiment-free model; all models except the weighting row use class-weighted training.}
\label{fig:effects}
\end{figure}

\begin{figure}[t]
\centering
\includegraphics[width=\columnwidth]{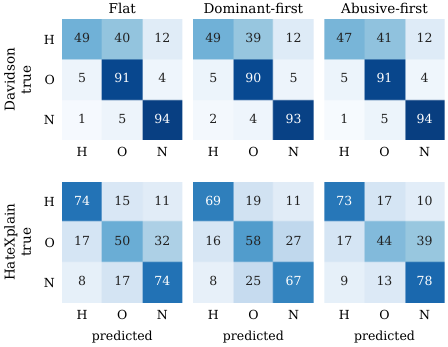}
\caption{Row-normalised confusion matrices (\%) for class-weighted LogReg without sentiment features. Davidson: pooled over the five test splits; HateXplain: official test split. Rows are true classes.}
\label{fig:confusion}
\end{figure}

The dominant-first hierarchy of HierSentNet was consistently worse than a flat classifier (\cref{tab:main,fig:effects}). With class weighting, macro-F1 fell by 1.1 points for LogReg ($p=0.007$) and 1.5 points for the SVM ($p<0.001$), and without class weighting by 1.8 and 1.2 points ($p<0.001$ and $p=0.011$). The abusive-first order performed on par with flat LogReg ($+0.1$, $p=0.79$) and slightly below the flat SVM ($-0.8$, $p=0.046$). The same ranking held on HateXplain, where the class-weighted dominant-first hierarchy was 1.4 points below flat for LogReg and 0.8 points below for the SVM.

\Cref{fig:confusion} shows where the loss comes from. On Davidson, hate recall was unchanged (48.5\% in both cases), so the dominant-first hierarchy did not lose more hateful tweets to the offensive class. Instead, the neither class became noisier: neither tweets labelled hate doubled (23 to 48 across the five test sets) and offensive tweets labelled neither rose from 421 to 458. Stage 2 is trained only on hate and neither, so every offensive tweet that slips past stage 1 is forced into one of those two classes. On HateXplain, where hate and offensive posts are more balanced, the hierarchy instead lost hate recall (73.9\%$\to$69.4\%) and normal recall (74.3\%$\to$67.3\%), both mostly to the offensive class. In both datasets, the first split groups two classes that are not similar and then removes the evidence needed to recover from its own mistakes.

\subsection{Do Sentiment Features Help? (RQ2)}

\begin{figure}[t]
\centering
\includegraphics[width=\columnwidth]{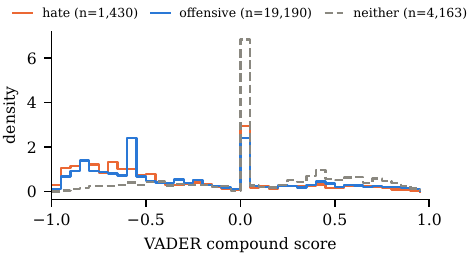}
\caption{Distribution of VADER compound scores by class on Davidson. Hateful and offensive tweets have almost the same sentiment profile; only the neither class differs.}
\label{fig:vader}
\end{figure}

\begin{table}[t]
\caption{VADER sentiment by Davidson class. Negative: compound $<-0.05$.}
\label{tab:vader}
\centering
\footnotesize
\setlength{\tabcolsep}{4pt}
\begin{tabular}{@{}lccc@{}}
\toprule
Class & Mean compound & Mean neg.\ score & Share negative \\
\midrule
Hate & $-0.35$ & 0.27 & 66.8\% \\
Offensive & $-0.29$ & 0.25 & 65.1\% \\
Neither & $+0.09$ & 0.07 & 24.9\% \\
\bottomrule
\end{tabular}
\end{table}

Adding sentiment features did not change performance in any class-weighted setting: the change in macro-F1 ranged from $-0.6$ to $0.0$ points, and no difference in hate-class F1 was significant (all $p\ge0.15$; \cref{fig:effects}). Without class weighting, the only significant effect was $+0.4$ macro-F1 for the flat SVM ($p=0.03$), with no significant change in its hate F1. On HateXplain, sentiment raised macro-F1 of the class-weighted flat LogReg from 66.2 to 67.0 and of the abusive-first LogReg from 65.0 to 66.8, but left the other class-weighted configurations within $\pm0.7$ points.

\Cref{fig:vader} and \cref{tab:vader} show why. Sentiment separates abusive from non-abusive language, which the text features already capture, but not hate from offence: the mean VADER compound score is $-0.35$ for hateful tweets and $-0.29$ for offensive tweets, and 66.8\% and 65.1\% of them are negative, against $+0.09$ and 24.9\% for neither. A signal that cannot separate the two abusive classes cannot fix the error that matters most, which is the confusion between hate and offence visible in \cref{fig:confusion} (40\% of Davidson hate tweets predicted offensive). This finding does not rule out richer affective models, such as emotion-category or target-aware multi-task training \citep{rajamanickam2020joint,plazadelarco2021multitask,awal2021angrybert}, but it shows that polarity alone is too coarse.

\subsection{Class Imbalance Matters Most (RQ3)}

Class-weighted training improved every Davidson configuration (\cref{tab:main,fig:effects}). For the flat models, macro-F1 rose by 2.8 points (LogReg) and 3.4 points (SVM), and hate-class F1 by 8.6 and 9.1 points (all $p<0.001$), mainly because hate recall nearly doubled (LogReg 26.6\%$\to$48.5\%). The cost was more offensive tweets misclassified as hate (1.5\%$\to$5.0\% for LogReg). Across all six decompositions the gain was 2.0--3.5 macro-F1 and 5.9--9.1 hate F1, several times larger than any architectural effect we measured, so a fair comparison of hierarchical or affect-aware models must control for class weighting first \citep{he2009learning,cui2019classbalanced}. On HateXplain, where classes are less imbalanced, weighting mainly traded hate F1 for offensive F1, with macro-F1 changing by at most 1.8 points.

The lexicon baseline reached a hate F1 of 45.2 on Davidson, on par with every learned model, but 52.9\% of the tweets it flagged were not hateful, confirming that keyword lists conflate hate with profanity \citep{kwok2013locate,davidson2017automated}.

\subsection{Robustness Across Datasets and Dialects (RQ4)}

\begin{table}[t]
\caption{Cross-dataset transfer (\%), LogReg with class weighting. D $\to$ H: train on Davidson, test on HateXplain; H $\to$ D: the reverse (seed-0 Davidson test split). In-domain flat macro-F1 is 74.5 (Davidson) and 66.2 (HateXplain).}
\label{tab:cross}
\centering
\footnotesize
\setlength{\tabcolsep}{2.2pt}
\resizebox{\columnwidth}{!}{%
\begin{tabular}{@{}llccccccc@{}}
\toprule
 & Decomp. & Sent. & Macro-F1 & Hate P & Hate R & Hate F1 & Off.\ F1 & Neither F1 \\
\midrule
D $\to$ H & Flat & -- & 45.1 & 45.9 & 56.7 & 50.7 & 35.4 & 49.3 \\
D $\to$ H & Flat & \checkmark & 44.6 & 46.4 & 52.0 & 49.0 & 38.5 & 46.3 \\
D $\to$ H & Dom.-first & -- & 44.4 & 42.9 & 57.6 & 49.2 & 36.0 & 47.9 \\
D $\to$ H & Dom.-first & \checkmark & 42.3 & 40.4 & 57.9 & 47.6 & 33.3 & 46.1 \\
D $\to$ H & Abus.-first & -- & 43.8 & 47.3 & 50.0 & 48.6 & 36.8 & 45.9 \\
D $\to$ H & Abus.-first & \checkmark & 43.4 & 45.4 & 52.9 & 48.9 & 34.9 & 46.4 \\
\midrule
H $\to$ D & Flat & -- & 50.7 & 25.0 & 19.6 & 22.0 & 80.5 & 49.5 \\
H $\to$ D & Flat & \checkmark & 50.1 & 25.7 & 18.9 & 21.8 & 79.5 & 49.1 \\
H $\to$ D & Dom.-first & -- & 53.2 & 24.2 & 16.8 & 19.8 & 85.5 & 54.2 \\
H $\to$ D & Dom.-first & \checkmark & 53.4 & 25.6 & 16.1 & 19.7 & 85.6 & 54.8 \\
H $\to$ D & Abus.-first & -- & 43.2 & 26.5 & 18.2 & 21.6 & 68.4 & 39.7 \\
H $\to$ D & Abus.-first & \checkmark & 43.0 & 27.4 & 18.2 & 21.8 & 67.4 & 39.7 \\
\bottomrule
\end{tabular}}
\end{table}

\begin{table}[t]
\caption{Lexical bias probe on Davidson: share (\%) of non-hateful test tweets classified as hate, with and without the reclaimed slur, pooled over five test splits (class-weighted models).}
\label{tab:bias}
\centering
\footnotesize
\setlength{\tabcolsep}{3.5pt}
\begin{tabular}{@{}lccc@{}}
\toprule
Model & With term & Without term & Ratio \\
 & ($n=916$) & ($n=10{,}764$) & \\
\midrule
LogReg, flat & 9.0 & 3.9 & 2.3$\times$ \\
LogReg, dominant-first & 8.2 & 4.5 & 1.8$\times$ \\
Linear SVM, flat & 4.9 & 3.1 & 1.6$\times$ \\
Linear SVM, flat + sentiment & 4.3 & 3.0 & 1.4$\times$ \\
\bottomrule
\end{tabular}
\end{table}

\textbf{Cross-dataset transfer.} Training on one dataset and testing on the other cut macro-F1 to 43--53, 21--31 points below in-domain performance (\cref{tab:cross}), in line with earlier generalisation studies \citep{arango2019hate,swamy2019studying}. Transfer to Davidson was hardest for the hate class (F1 20--22): with the flat model, 83 of the 143 hateful test tweets were predicted offensive, and 66 offensive tweets were predicted hateful, so the two datasets draw the hate/offensive boundary in different places. No decomposition was robust in both directions: the dominant-first hierarchy was best for HateXplain$\to$Davidson (53.2) but only second in the other direction. Adding sentiment did not improve transfer; macro-F1 changed by $-2.1$ to $+0.2$ points.

\textbf{Lexical bias.} Of the non-hateful Davidson test tweets that contain the reclaimed slur (916 across the five test sets, all labelled offensive), the flat class-weighted LogReg labelled 9.0\% as hate, against 3.9\% of non-hateful tweets without it, a 2.3$\times$ higher false-positive rate (\cref{tab:bias}). The gap was smaller for the SVM (4.9\% vs.\ 3.1\%), and adding sentiment features lowered the SVM's rate to 4.3\%. In the full dataset, 1,742 of the 1,944 tweets containing this term are labelled offensive and 202 hateful, so a model can lower its loss by treating the term itself as evidence of hate. Because the term is common in African-American English, the gap is consistent with the dialect bias reported by \citet{sap2019risk} and \citet{davidson2019racial}.

\section{Extended Analyses (RQ5)}
\label{sec:extended}

The results above could depend on choices a critic might call weak: polarity is a crude affect signal, hard hierarchies propagate errors, class weighting is one of many imbalance remedies, and linear models cannot represent context. We test each of these in turn on the same splits.

\begin{figure*}[t]
\centering
\includegraphics[width=\textwidth]{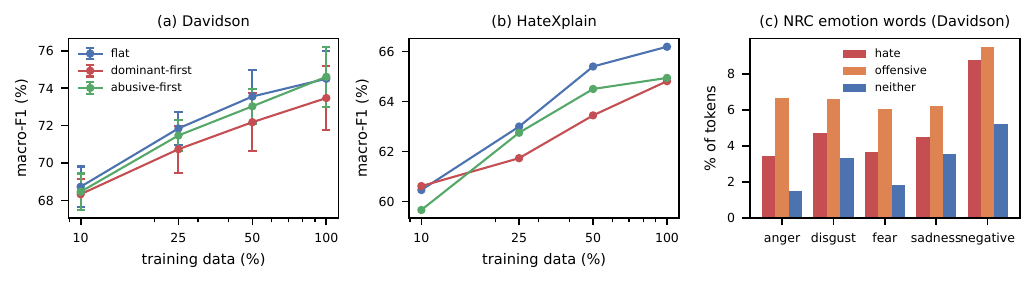}
\vspace{-6mm}
\caption{(a, b) Learning curves for class-weighted LogReg on Davidson (mean\,$\pm$\,std over five splits) and HateXplain: the flat classifier is never clearly worse than either hierarchy. (c) Share of tokens associated with each NRC emotion, by Davidson class. Offensive tweets contain more anger, disgust, fear and sadness words than hateful tweets, so emotion lexicons, like polarity, point in the wrong direction for the hate/offence boundary.}
\label{fig:extended}
\vspace{-2mm}
\end{figure*}

\begin{table}[t]
\caption{Polarity versus emotion features (class-weighted flat LogReg, \%). $\Delta$ and $p$: change in Davidson macro-F1 against no affect features, paired $t$-test over the five splits.}
\label{tab:affect}
\centering
\footnotesize
\setlength{\tabcolsep}{2.6pt}
\resizebox{\columnwidth}{!}{%
\begin{tabular}{@{}lcccccc|cc@{}}
\toprule
 & \multicolumn{6}{c|}{Davidson (5 splits)} & \multicolumn{2}{c}{HateXplain}\\
\cmidrule(lr){2-7}\cmidrule(l){8-9}
Variant & Macro-F1 & $\Delta$ & $p$ & Hate F1 & Hate R & \textsc{o}$\to$\textsc{h} & Macro-F1 & Hate F1\\
\midrule
No affect features & 74.5\sd{1.5} & -- & -- & 44.2 & 48.5 & 5.0 & 66.2 & 73.7\\
+ VADER polarity & 74.4\sd{1.2} & -0.1 & .84 & 44.4 & 50.3 & 5.3 & 67.0 & 74.9\\
+ NRC emotions & 74.4\sd{1.3} & -0.1 & .46 & 44.1 & 49.4 & 5.1 & 66.3 & 73.9\\
+ VADER + NRC & 74.6\sd{1.3} & +0.1 & .75 & 44.8 & 50.3 & 5.2 & 67.3 & 74.8\\
\bottomrule
\end{tabular}}
\end{table}

\textbf{Emotion categories do not help either.} Prior work reports gains from emotion categories rather than polarity \citep{rajamanickam2020joint,plazadelarco2021multitask}, so we replaced and combined the VADER features with NRC emotion features. On Davidson, no variant changed macro-F1 by more than 0.13 points (all $p\ge0.45$) or hate F1 by more than 0.6 points (\cref{tab:affect}). \Cref{fig:extended}c explains why: offensive tweets contain \emph{more} emotion words than hateful tweets (6.7\% vs.\ 3.5\% of tokens for anger and 6.6\% vs.\ 4.7\% for disgust), because the lexicon responds to the profanity and insults that define the offensive class. On HateXplain, VADER plus NRC features raised macro-F1 from 66.2 to 67.3 and hate F1 from 73.7 to 74.8, the largest affect effect we observed but still on a single split. Lexicon-level affect therefore marks abusive language, not hateful language.

\begin{table}[t]
\caption{Hard versus probabilistic (soft) hierarchical inference (class-weighted LogReg, \%). Soft inference multiplies the stage probabilities and takes the most probable class. $\Delta$, $p$: against flat.}
\label{tab:soft}
\centering
\footnotesize
\setlength{\tabcolsep}{2.6pt}
\resizebox{\columnwidth}{!}{%
\begin{tabular}{@{}lcccccc|cc@{}}
\toprule
 & \multicolumn{6}{c|}{Davidson (5 splits)} & \multicolumn{2}{c}{HateXplain}\\
\cmidrule(lr){2-7}\cmidrule(l){8-9}
Variant & Macro-F1 & $\Delta$ & $p$ & Hate F1 & Hate R & \textsc{o}$\to$\textsc{h} & Macro-F1 & Hate F1\\
\midrule
Flat & 74.5\sd{1.5} & -- & -- & 44.2 & 48.5 & 5.0 & 66.2 & 73.7\\
Dom.-first, hard & 73.5\sd{1.7} & -1.0 & .07 & 42.6 & 48.5 & 5.3 & 64.8 & 71.4\\
Dom.-first, soft & 73.7\sd{1.7} & -0.8 & .15 & 42.5 & 45.5 & 4.6 & 63.8 & 71.0\\
Abus.-first, hard & 74.6\sd{1.6} & +0.1 & .75 & 42.9 & 46.7 & 5.1 & 64.9 & 72.7\\
Abus.-first, soft & 74.0\sd{1.2} & -0.5 & .25 & 41.8 & 44.1 & 4.8 & 66.1 & 75.2\\
\bottomrule
\end{tabular}}
\end{table}

\textbf{The problem is the decomposition, not hard decisions.} If the dominant-first hierarchy failed only because stage-1 errors cannot be undone, probabilistic inference should close the gap. It does not (\cref{tab:soft}): soft inference moved dominant-first macro-F1 from 73.5 to 73.7 on Davidson, still 0.8 points below flat, and lowered it on HateXplain (64.8 to 63.8). Soft inference helped the abusive-first order on HateXplain (macro-F1 66.1 and hate F1 75.2, against 73.7 for the class-weighted flat model) but lowered its Davidson hate F1 by 2.4 points relative to flat ($p=0.026$). No hierarchical variant beat the flat classifier on Davidson by more than 0.1 macro-F1 points.

\textbf{Hierarchies do not become useful with less data.} One argument for decomposition is that simpler sub-problems need less data. The learning curves (\cref{fig:extended}a,b) do not support this. On Davidson, the flat classifier matched or beat both hierarchies at every training size (10\%: 68.7 vs.\ 68.3 and 68.5; 50\%: 73.6 vs.\ 72.2 and 73.0). On HateXplain, the dominant-first hierarchy was ahead of flat only at 10\% of the data, by 0.2 points, and fell behind as data grew.

\begin{table}[t]
\caption{Strategies for class imbalance (flat LogReg, \%). Decision bias adds per-class offsets to the log-probabilities, tuned on validation macro-F1. $\Delta$, $p$: against no correction.}
\label{tab:imb}
\centering
\footnotesize
\setlength{\tabcolsep}{2.6pt}
\resizebox{\columnwidth}{!}{%
\begin{tabular}{@{}lcccccc|cc@{}}
\toprule
 & \multicolumn{6}{c|}{Davidson (5 splits)} & \multicolumn{2}{c}{HateXplain}\\
\cmidrule(lr){2-7}\cmidrule(l){8-9}
Variant & Macro-F1 & $\Delta$ & $p$ & Hate F1 & Hate R & \textsc{o}$\to$\textsc{h} & Macro-F1 & Hate F1\\
\midrule
None & 71.7\sd{2.0} & -- & -- & 35.7 & 26.6 & 1.5 & 65.9 & 75.7\\
Class weights & 74.5\sd{1.5} & +2.8 & .03 & 44.2 & 48.5 & 5.0 & 66.2 & 73.7\\
Random oversampling & 74.0\sd{1.3} & +2.3 & .05 & 43.7 & 51.7 & 5.8 & 65.8 & 73.7\\
Random undersampling & 68.7\sd{0.5} & -3.0 & .02 & 37.2 & 70.9 & 14.8 & 65.6 & 73.5\\
Decision bias (val.) & 74.8\sd{1.1} & +3.1 & .01 & 44.4 & 51.7 & 5.5 & 66.4 & 74.8\\
Weights + bias & 74.5\sd{1.3} & +2.8 & .010 & 44.0 & 50.8 & 5.3 & 66.2 & 73.7\\
\bottomrule
\end{tabular}}
\end{table}

\textbf{Any reasonable imbalance remedy works; undersampling does not.} \Cref{tab:imb} compares five remedies against an uncorrected flat LogReg. Class weights, oversampling, a validation-tuned decision bias and weights plus bias all raised Davidson macro-F1 by 2.3--3.1 points and hate F1 by 8.0--8.7 points (all $p\le0.05$), and differed from each other by less than one seed standard deviation. The cheapest option, a decision bias added after training, was marginally the best (74.8). Random undersampling raised hate recall to 70.9\% but labelled 14.8\% of offensive tweets as hate and lowered macro-F1 by 3.0 points, because it discards most of the training data. On HateXplain, where hate is not a minority class, all remedies stayed within 0.5 macro-F1 points of no correction, and class weighting \emph{lowered} hate F1 from 75.7 to 73.7. Imbalance handling helps when, and only when, the target class is rare.

\textbf{Class weighting harms calibration.} The uncorrected Davidson model was well calibrated (expected calibration error 0.019 on average, range 0.014--0.024), whereas the class-weighted model's error averaged 0.054 and reached 0.096 on one split; on HateXplain it rose from 0.058 to 0.150. Weighting shifts predicted probabilities towards the minority classes, so its scores should not be used as probabilities, for instance as a moderation threshold, without recalibration. A post-hoc decision bias achieves the same F1 while leaving the model's probabilities untouched.

\begin{table}[t]
\caption{TextCNN trained from scratch (\%). Davidson: mean\,$\pm$\,std over five splits; $\Delta$ and $p$ against the class-weighted flat CNN (paired $t$-test). HateXplain: official split, one run.}
\label{tab:cnn}
\centering
\footnotesize
\setlength{\tabcolsep}{2.6pt}
\resizebox{\columnwidth}{!}{%
\begin{tabular}{@{}lccccc|cc@{}}
\toprule
 & \multicolumn{5}{c|}{Davidson (5 splits)} & \multicolumn{2}{c}{HateXplain}\\
\cmidrule(lr){2-6}\cmidrule(l){7-8}
Variant & Macro-F1 & $\Delta$ & $p$ & Hate F1 & Hate R & Macro-F1 & Hate F1\\
\midrule
Flat & 69.0\sd{2.7} & -3.1 & .05 & 28.3 & 20.1 & 61.5 & 71.2\\
Flat, weighted & 72.1\sd{1.0} & -- & -- & 41.3 & 53.3 & 63.5 & 71.6\\
Flat, weighted + VADER & 72.3\sd{1.6} & +0.3 & .64 & 40.9 & 51.7 & 62.2 & 69.0\\
Dom.-first, weighted & 70.4\sd{3.0} & -1.6 & .24 & 37.6 & 40.0 & 62.7 & 70.3\\
Abus.-first, weighted & 71.4\sd{0.7} & -0.6 & .23 & 37.4 & 51.9 & 62.0 & 66.3\\
\bottomrule
\end{tabular}}
\end{table}

\textbf{A neural model gives the same picture.} The TextCNN learns its own representation instead of using fixed n-grams, yet every design effect kept its direction (\cref{tab:cnn}). Class weighting was again the largest lever: +3.1 macro-F1 ($p=0.048$) and +13.0 hate F1 ($p=0.018$), with hate recall rising from 20.1\% to 53.3\%. VADER features changed macro-F1 by +0.3 points ($p=0.64$) and hate F1 by $-0.3$. Both hierarchies were below the class-weighted flat CNN: the dominant-first order by 1.6 macro-F1 and 3.7 hate F1 points, the abusive-first order by 0.6 and 3.9 points (hate F1 $p=0.026$). The same ordering held on HateXplain (flat 63.5, dominant-first 62.7, abusive-first 62.0, flat with VADER 62.2). Trained from scratch on 20k tweets, the CNN was 2.4 macro-F1 points below the linear model, as expected without pretraining, but this does not change the comparison between designs.

\begin{table}[t]
\caption{Hate recall by target community on the HateXplain test set (\%, flat LogReg). ``Abusive counts a hateful post as caught if it is predicted hate or offensive. Groups with at least 20 hateful test posts.}
\label{tab:target}
\centering
\footnotesize
\setlength{\tabcolsep}{4pt}
\begin{tabular}{@{}lrccc@{}}
\toprule
Target & $n$ & Weighted & Unweighted & Abusive\\
\midrule
Jewish & 155 & 84.5 & 85.2 & 94.8\\
African & 261 & 83.5 & 85.4 & 92.7\\
Islam & 149 & 74.5 & 76.5 & 84.6\\
Hispanic & 32 & 71.9 & 75.0 & 87.5\\
Arab & 103 & 70.9 & 72.8 & 87.4\\
Asian & 26 & 61.5 & 65.4 & 76.9\\
Women & 99 & 60.6 & 67.7 & 84.8\\
Homosexual & 90 & 56.7 & 65.6 & 88.9\\
Refugee & 38 & 55.3 & 55.3 & 76.3\\
Caucasian & 39 & 53.8 & 56.4 & 74.4\\
\bottomrule
\end{tabular}
\end{table}

\textbf{Who is missed?} Hate recall varied widely across target communities on HateXplain (\cref{tab:target}): 84.5\% for hate against Jewish people and 83.5\% against African people, but 60.6\% for women, 56.7\% for homosexual people, 55.3\% for refugees and 53.8\% for Caucasian people. Most missed posts were labelled offensive rather than normal: counting a prediction of hate or offensive as a catch, recall was 74--95\% for every group. The model is therefore much better at recognising hate that uses group-specific slurs than hate expressed through gendered insults or generic language, which aggregate scores hide. Class weighting lowered or left unchanged the recall of every group here, consistent with the HateXplain imbalance results above.

\section{Further Benchmarks and Diagnostics (RQ6)}
\label{sec:bench}

The study so far uses two datasets whose label sets are flat. We now test the design choices on a dataset annotated hierarchically, probe the detectors with functional tests, and examine where in-domain errors come from.

\begin{table}[t]
\caption{OLID \citep{zampieri2019predicting}, official test set (860 tweets), class-weighted LogReg, mean\,$\pm$\,std over five validation splits (\%). Fine labels: \textsc{not}, untargeted (\textsc{unt}), and targeted at an individual (\textsc{ind}), a group (\textsc{grp}) or other (\textsc{oth}). Level A: offensive vs.\ not; level B: targeted vs.\ untargeted among offensive tweets; level C: target type among targeted tweets. $p$: paired $t$-test against the row above.}
\label{tab:olid}
\centering
\footnotesize
\setlength{\tabcolsep}{2.4pt}
\resizebox{\columnwidth}{!}{%
\begin{tabular}{@{}llcccccc@{}}
\toprule
Sent. & Decomposition & 5-way F1 & $p$ & Level A & $p$ & Level B & Level C\\
\midrule
-- & Flat & 44.9\sd{2.2} & -- & 72.8\sd{0.8} & -- & 46.0 & 45.4\\
-- & A$\to$B$\to$C & 44.6\sd{0.4} & .80 & 74.2\sd{0.6} & .009 & 43.4 & 44.4\\
\midrule
\checkmark & Flat & 47.7\sd{1.6} & -- & 75.3\sd{0.7} & -- & 47.8 & 47.3\\
\checkmark & A$\to$B$\to$C & 46.8\sd{2.0} & .38 & 75.2\sd{1.7} & .880 & 45.6 & 45.8\\
\bottomrule
\end{tabular}}
\end{table}

\textbf{A hierarchy that matches the annotation helps its first decision only.} OLID \citep{zampieri2019predicting} was annotated top-down: offensive or not (level A), then targeted or untargeted (B), then the type of target (C), so a classifier hierarchy that follows these levels matches how the labels were produced. We train on the 13,240 official training tweets (10\% held out for validation, five seeds) and test on the official 860. Without sentiment features, the A$\to$B$\to$C hierarchy improved level-A macro-F1 from 72.8 to 74.2 ($p=0.009$) and reduced the seed variance of the five-way score from 2.2 to 0.4 points, but the five-way macro-F1 did not change (44.6 vs.\ 44.9, $p=0.80$), because levels B and C fell by 2.6 and 1.0 points (\cref{tab:olid}). Group-targeted tweets, the OLID class closest to hate speech, reached the same F1 with and without the hierarchy (40.6 vs.\ 40.4). Together with the Davidson and HateXplain results, this supports a narrower conclusion than ``hierarchies help'': a coarse-to-fine split helps the coarse decision when it follows the annotation, and hurts when it groups classes that are not similar, as the dominant-first split does.

\textbf{Sentiment helps where it separates the classes.} On OLID, VADER features raised flat five-way macro-F1 by 2.8 points ($p=0.005$) and level-A macro-F1 by 2.5 points ($p=0.008$), the largest sentiment effect in this study. This is consistent with \cref{tab:vader}: polarity separates abusive from non-abusive text, and OLID's main decision is exactly that, whereas the hate/offensive boundary in Davidson and HateXplain is not. With sentiment features the hierarchy no longer helped level A (75.2 vs.\ 75.3), because sentiment already supplied the information the first stage added.

\begin{table}[t]
\caption{HateCheck functional tests \citep{rottger2021hatecheck} for class-weighted LogReg trained on Davidson (seed-0 split) or HateXplain: accuracy (\%) per group of functionalities, weighted by number of cases. A case counts as correct if a hateful case is predicted \textsc{h} or a non-hateful case is not.}
\label{tab:hatecheck}
\centering
\footnotesize
\setlength{\tabcolsep}{2.6pt}
\resizebox{\columnwidth}{!}{%
\begin{tabular}{@{}lrcccc@{}}
\toprule
 & & \multicolumn{3}{c}{Trained on Davidson} & HateXplain\\
\cmidrule(lr){3-5}\cmidrule(l){6-6}
Functionality group & $n$ & Flat & Dom.-first & Flat+Sent. & Flat\\
\midrule
\multicolumn{6}{@{}l}{\emph{Hateful cases}}\\
Derogation & 560 & 30.5 & 39.8 & 34.6 & 8.2\\
Threats & 273 & 32.2 & 46.2 & 37.0 & 12.5\\
Slurs & 144 & 48.6 & 47.2 & 46.5 & 33.3\\
Hateful profanity & 140 & 36.4 & 44.3 & 45.7 & 27.9\\
Pronoun reference & 273 & 29.3 & 50.9 & 32.6 & 11.7\\
Phrasing (question/opinion) & 273 & 27.1 & 42.1 & 32.6 & 8.1\\
Negated positive statement & 140 & 20.7 & 35.7 & 20.7 & 7.1\\
Spelling variations & 760 & 30.7 & 38.7 & 24.3 & 9.6\\
\multicolumn{6}{@{}l}{\emph{Non-hateful cases}}\\
Non-hateful profanity & 100 & 100.0 & 95.0 & 100.0 & 82.0\\
Reclaimed slurs, homonyms & 111 & 64.0 & 37.8 & 56.8 & 92.8\\
Negated hateful statement & 133 & 72.2 & 62.4 & 69.2 & 87.2\\
Neutral/positive identity & 315 & 80.3 & 72.4 & 84.1 & 97.8\\
Counter-speech & 314 & 58.6 & 49.7 & 54.5 & 86.9\\
Abuse of non-protected targets & 192 & 91.7 & 82.8 & 85.4 & 86.5\\
\midrule
All hateful & 2563 & 31.1 & 42.0 & 31.9 & 11.9\\
All non-hateful & 1165 & 75.5 & 65.5 & 73.4 & 90.0\\
Overall & 3728 & 45.0 & 49.4 & 44.9 & 36.3\\
\bottomrule
\end{tabular}}
\end{table}

\textbf{Functional tests expose low recall.} HateCheck \citep{rottger2021hatecheck} contains 3,728 short template cases in 29 functionalities. The class-weighted flat model trained on Davidson labelled only 31.1\% of the 2,563 hateful cases as hate, and the model trained on HateXplain only 11.9\% (\cref{tab:hatecheck}). Recall was lowest for negated positive statements (20.7\% and 7.1\%) and highest for slurs (48.6\% and 33.3\%), so both detectors rely on lexical cues rather than on what a sentence says about a group. The two models fail in opposite directions on non-hateful cases: the Davidson model labelled 36\% of reclaimed slurs and homonyms and 41\% of counter-speech as hate, whereas the HateXplain model was correct on 87--98\% of these cases but missed most hate. The dominant-first hierarchy raised Davidson hate recall to 42.0\% while lowering non-hateful accuracy from 75.5\% to 65.5\%, which is a shift of the decision threshold rather than better understanding. Sentiment features changed overall accuracy by $-0.1$ points. In-domain macro-F1 of 66--75 therefore coexists with functional accuracy below 50\%.

\begin{table}[t]
\caption{Diagnostics (\%). Top: representation ablation for class-weighted flat LogReg on Davidson (five splits; $p$ against word + char). Middle: HateXplain test posts by annotator agreement. Bottom: expected calibration error (ECE) of class-weighted LogReg before and after temperature scaling fitted on validation data.}
\label{tab:diag}
\centering
\footnotesize
\setlength{\tabcolsep}{3pt}
\begin{tabular}{@{}lccc@{}}
\toprule
Representation & Macro-F1 & $p$ & Hate F1\\
\midrule
word 1--2 & 74.3\sd{1.0} & .672 & 44.7\\
char 2--5 & 72.8\sd{0.8} & .008 & 41.5\\
word + char & 74.5\sd{1.5} & -- & 44.2\\
\midrule
Annotator agreement & Macro-F1 & $n$ & Hate F1\\
\midrule
3/3 majority & 75.6 & 974 & 85.4\\
2/3 majority & 55.3 & 950 & 61.4\\
\midrule
Calibration & ECE before & ECE after & $T$\\
\midrule
Davidson (mean of 5) & 5.4 & 1.9 & 0.87\\
HateXplain & 15.0 & 3.0 & 0.52\\
\bottomrule
\end{tabular}
\end{table}

\textbf{Disagreement bounds performance.} HateXplain releases all three annotations, so test posts can be split by agreement \citep{davani2022dealing,uma2021learning}. On the 974 unanimous posts, flat LogReg reached 75.6 macro-F1 and 85.4 hate F1; on the 950 posts with a 2/3 majority, 55.3 and 61.4 (\cref{tab:diag}). Of the 628 HateXplain test errors, 427 (68\%) are on posts that one of three annotators labelled differently, which limits how much any model can gain from better features and argues for reporting results by agreement level.

\textbf{Word n-grams carry most of the signal.} Word 1--2-grams alone matched the combined representation (74.3 vs.\ 74.5 macro-F1, $p=0.67$), while character 2--5-grams alone were 1.7 points lower ($p=0.008$). Character features help with spelling variants, which are rare in Davidson but form 20\% of HateCheck, where the Davidson model still caught only 30.7\% of misspelt hateful cases.

\textbf{Temperature scaling repairs calibration.} \Cref{sec:extended} showed that class weighting harms calibration. A single temperature fitted on validation data, which does not change predictions, lowered the expected calibration error from 5.4\% to 1.9\% on Davidson and from 15.0\% to 3.0\% on HateXplain (\cref{tab:diag}). Class-weighted training plus temperature scaling therefore gives both the recall gain of weighting and usable probabilities.

\section{Discussion and Limitations}
\label{sec:discussion}

\textbf{What the evidence supports.} (i) A hierarchy only helps if its first split matches the structure of the data: on OLID, whose annotation is hierarchical, it improved the first decision, whereas the dominant-class-first split of HierSentNet hurts because its second stage never sees the dominant class, and neither probabilistic inference nor smaller training sets change this. (ii) Lexicon-level affect, whether polarity or emotion categories, cannot separate hate from offence, because offensive language is at least as negative and emotional as hateful language; it does help to separate offensive from non-offensive text, as on OLID. (iii) Handling class imbalance is the largest single lever when the hate class is rare, but the specific remedy matters little, and a post-hoc decision bias or temperature scaling avoids its damage to calibration. (iv) In-domain scores overstate robustness by 21--31 macro-F1 points across datasets and hide a lexical bias, large differences in recall between target communities and functional accuracy below 50\% on HateCheck. (v) All four findings hold for a convolutional network as well as for linear models.

\textbf{Why linear models.} Linear n-gram models make the comparison cheap enough to run 24 configurations over five splits and two datasets with significance tests, and they remain competitive baselines on these datasets. Their weakness is that they cannot represent context, so the absolute numbers are lower than those of fine-tuned transformers. The direction of the design effects is what we test here, and the convolutional network, which learns its own representation, shows the same directions.

\textbf{Limitations.} Pretrained transformers may use sentiment differently, so running the released script on a GPU is needed before our conclusions can be extended to them. We test lexicon-level polarity and emotion features, not multi-task emotion learning, which learns affect from labelled data rather than from a word list. All three datasets are English and drawn from Twitter and Gab, their labels reflect specific annotator pools and definitions of hate, and majority-vote labels hide disagreement \citep{davani2022dealing}, which bounds in-domain scores (\cref{tab:diag}). HateCheck cases are short templates, so its scores measure specific capabilities rather than performance on real posts. The bias probe uses a single term as a proxy for dialect and is not a full fairness audit. Finally, we report one HateXplain split, so its differences below one point should not be over-interpreted, and the paired $t$-tests over five Davidson splits have limited power for differences below about one point.

\section{Conclusion}
\label{sec:conclusion}

We evaluated HierSentNet, a hierarchical, sentiment-augmented framework for hate speech detection, component by component with a controlled protocol. On two benchmarks, the dominant-first hierarchy reduced macro-F1 by 0.8--1.5 points, sentiment features changed Davidson macro-F1 by at most 0.6 points because hateful and offensive posts are almost equally negative, and class-weighted training improved flat-model hate F1 by 8.6--9.1 points. Detectors trained on one dataset lost 21--31 macro-F1 points on the other and over-flagged a reclaimed slur. These conclusions held for emotion features, probabilistic hierarchies, smaller training sets, other imbalance remedies and a convolutional network. On hierarchically annotated OLID, a matching hierarchy improved only the first decision and sentiment helped, which shows when each idea can work. Functional tests showed that the detectors caught at most 31\% of hateful HateCheck cases. Future work should re-run the transformer variants on the released splits, test multi-task emotion learning, and train with disagreement-aware objectives.

\section*{Impact Statement}

This work aims to improve the reliability of automated hate speech detection, which can reduce exposure to harmful content online. Such systems also carry risks: false positives can silence legitimate speech, and we show that one detector flags tweets containing a term common in African-American English at more than twice the base rate, which could disproportionately affect Black users. Detectors trained on one dataset also degrade sharply on another, and on functional tests they missed most hateful statements. Recall also differs by target community: on HateXplain, hate against women, homosexual people and refugees was missed far more often than hate against Jewish or African people, so these groups would receive less protection from automated moderation. We therefore recommend that these models support human moderators rather than act autonomously, and that deployments audit error rates across dialects and communities. The datasets used contain offensive and hateful text; we quote none of it in the paper.


\bibliographystyle{icml2026}

\end{document}